%% file: 00_main.tex
\documentclass[11pt,a4paper]{article}
\usepackage[hyperref]{emnlp2020}
\usepackage{times}
\usepackage{latexsym}

\usepackage{trackchanges}
\addeditor{YJ}

\newcommand\shortsection[1]{\vspace{6pt}{\noindent\bf #1.}}

\usepackage{color}

\newcommand{\ynote}[1]{\textcolor{red}{\textbf{Yangfeng:} #1}}

\usepackage{microtype}
\usepackage[export]{adjustbox}
\usepackage{subcaption}
\usepackage{booktabs}
\usepackage{amssymb}
\usepackage{stfloats}
\usepackage{multirow}
\usepackage{tabularx}
\usepackage{csvsimple}
\usepackage{longtable}
\usepackage{hyperref}
\usepackage{array}

\usepackage{caption}

\aclfinalcopy 

\newcommand{\flipped}[1]{\dataset{#1-Flipped}} 
\newcommand{\dataset}[1]{\textsf{\small #1}}

\title{Finding Friends and Flipping Frenemies: \\ Automatic Paraphrase Dataset Augmentation Using Graph Theory}

\author{Hannah Chen, 
    Yangfeng Ji, David Evans \\
    Department of Computer Science\\
    University of Virginia\\
    Charlottesville, VA 22904\\
  \texttt{\{yc4dx,yangfeng,evans\}@virginia.edu} \\}

\date{}

\begin{document}

\maketitle

\input{0_abstract}

\input{1_introduction}
\input{2_background}
\input{3_method}
\input{4_experiments}

\input{6_related}

\input{7_conclusion}


\bibliographystyle{acl_natbib_edit}
\bibliography{anthology,emnlp2020}

\clearpage
\input{appendix}

\end{document}

%% file: 0_abstract.tex
\begin{abstract}
Most NLP datasets are manually labeled, so suffer from inconsistent labeling or limited size. We propose methods for automatically improving datasets by viewing them as graphs with expected semantic properties. We construct a paraphrase graph from the provided sentence pair labels, and create an augmented dataset by directly inferring labels from the original sentence pairs using a transitivity property. We use structural balance theory to identify likely mislabelings in the graph, and flip their labels. We evaluate our methods on paraphrase models trained using these datasets starting from a pretrained BERT model, and find that the automatically-enhanced training sets result in more accurate models.
\end{abstract}

%% file: 1_introduction.tex
\section{Introduction}
Having high quality annotated data is crucial for training supervised machine learning models. However, producing large datasets with good labeling quality is expensive and labor intensive. Most NLP datasets rely on labels provided by 
human annotators with varying skills and limited training and expertise. The label instances are also often based on ambiguous definitions and guidelines. 

To address this problem, we study automated techniques to improve datasets for training and testing. In particular, we focus on paraphrase identification task, which aims to determine whether two given sentences are semantically equivalent. The sentences and labels in a dataset can be viewed as nodes and edges of a graph. Moving from single labeled sentence pairs to a graph provides a better understanding of the sentence relations of the dataset, which can be exploited to infer additional edge labels. In particular, since paraphrases are an equality relation, we can perform a transitive closure on the graph to infer additional labels.
In addition, we use the notion of balance~\citep{harary1953} for signed graphs to identify conflicted relations. In the context of semantic relationships between pairs of sentences, any paraphrases of a given sentence cannot be a non-paraphrase of each other since they should all share an identical meaning.

\shortsection{Contributions} 
We show the benefits of representing sentence pair relations as a graph. We first construct a paraphrase graph with the original pairs and their relation labels from the Quora Question Pairs (QQP) dataset~\citep{qqp} following the structure of a signed graph. With the graph structure and the
transitivity of paraphrases, we can automatically infer new sentence pair relations directly from the original dataset (Section~\ref{infer-labels}). In addition, we identify and correct likely mislabeled pairs based on violations of expected structural balance properties we expect a valid paraphrase graph to satisfy 
(Section~\ref{identify-mislabeling}). We found 90 seemingly mislabeled sentence pairs in the QQP dataset.  We show that fine-tuning a BERT model on the augmented set improves its performance on both the original and augmented testing sets, decreasing the error rate from 10\% to under 6\% when testing on the augmented test set. 
We released the augmented QQP dataset and the implementation code. (\href{https://github.com/hannahxchen/automatic-paraphrase-dataset-augmentation}{https://github.com/hannahxchen/automatic-paraphrase-dataset-augmentation})

%% file: 2_background.tex
\section{Representing Datasets as Graphs}

A \emph{signed graph} is a graph where each edge is labeled either positive or negative to indicate a relationship between the two connected nodes. For undirected graphs, this relationship is symmetric. A path is a set of connected edges with no repeated nodes, and a path with the last node connecting back to the first node forms a cycle.  Given semantic interpretations of the edge labels, all paths in a signed graph should have certain properties.

\begin{figure}[tb]
\centering
\begin{subfigure}[b]{0.2\textwidth}
\includegraphics[width=0.7\linewidth, center]{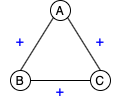} 
\caption{}
\end{subfigure}
\vspace{2mm}
\begin{subfigure}[b]{0.2\textwidth}
\includegraphics[width=0.7\linewidth, center]{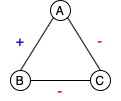}
\caption{}
\end{subfigure}
\begin{subfigure}[b]{0.2\textwidth}
\includegraphics[width=0.7\linewidth, center]{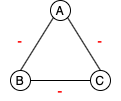}
\caption{}
\label{weakly-balanced}
\end{subfigure}
\begin{subfigure}[b]{0.2\textwidth}
\includegraphics[width=0.7\linewidth, center]{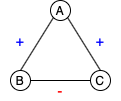}
\caption{}
\end{subfigure}

\caption{Four types of signed graphs of a triad. Signed graphs (a) and (b) are balanced; (c) is weakly balanced; (d) is imbalanced.}
\label{structral-balance}
\end{figure}

\shortsection{Structural Balance}
Balance theory was proposed by \citet{heider-1946} to study interpersonal relationships in social psychology. The idea was generalized to signed graphs by \citet{harary1953}. A graph is said to be \emph{balanced} if the product of the edge signs in every cycle is positive. There are only two types of conditions exist in a balanced signed graph: (1) all the nodes are connected with only positive edges, or (2) nodes can be divided into subsets such that nodes within each subset are connected with positive edges and nodes from different subsets are connected with negative edges. Figure~\ref{structral-balance} illustrates four possible sign combinations for a triad.

\begin{figure*}[tb]
\centering
\includegraphics[width=1\linewidth]{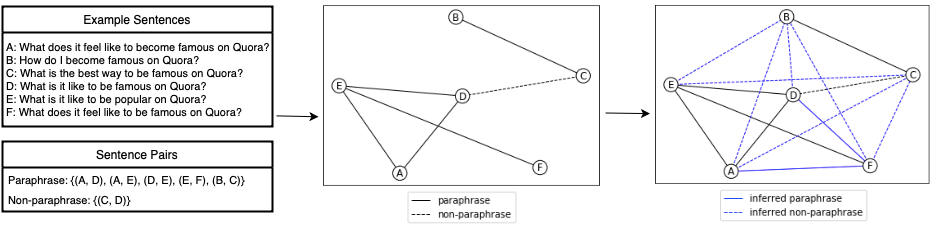}
\caption{Example paraphrase graph constructed from example pairs from the Quora Question Pairs (QQP) dataset. The right most figure shows the inferred relations from the paraphrase graph.}
\label{paraphrase-graph}
\end{figure*}

\shortsection{Paraphrase Graph}
The definition of a paraphrase remains ambiguous and varies by task, but the most common definitions are similar to the one from \citet{bhagat-hovy-2013-squibs}, which define paraphrases as sentences that convey the same meaning but are expressed in different forms. Since this notion is a symmetric relation, we can form an undirected signed graph by linking the sentence pairs from the paraphrase dataset with their annotated relations. Sentence pairs labeled as paraphrases are connected with positive edges; sentences labeled as non-paraphrases are connected with negative edges. A paraphrase cluster contains sentences connected with positive edges, and all sentences in the cluster should share the same meaning.
Figure~\ref{paraphrase-graph} shows how a paraphrase graph is constructed from selected labeled pairs in the QQP dataset. 


%% file: 3_method.tex
\section{Improving Datasets using Graphs}
\label{methods}

Typically, training sets for paraphrase identification are constructed by using annotations for sentence pairs provided by human annotators. Based on the semantics implied by the paraphrase and non-paraphrase labels, we can augment and correct the sentence-level paraphrase graph. Our method infers labels based on transitivity (Section~\ref{sec:inferring}), and identifies likely mislabelings based on expected graph consistency properties (Section~\ref{sec:flipping}).

\subsection{Inferring New Labels (Finding Friends)}
\label{infer-labels}\label{sec:inferring}
Since paraphrase is a reflexive, symmetric, and transitive relation, we can identify a set of semantically equivalent sentences if they are reachable by one another along the paraphrase links. We use Dijkstra's shortest path algorithm~\citep{dijkstra1959} implemented by Networkx~\citep{Haberg2008Networkx} to find paraphrase paths between nodes. Furthermore, we can infer additional non-paraphrase edges between nodes from two different paraphrase clusters if they are connected with one or more non-paraphrase links. Figure~\ref{paraphrase-graph} illustrates how a paraphrase graph with inferred edge labels is constructed. For example, we can infer a positive link from node A to F, and a negative link from node A to C since A and D are paraphrases and C and D are non-paraphrases. By applying this method to the entire dataset, we expand the training set size for QQP by 60.7\% (Section ~\ref{datasets}).


\subsection{Fixing Mislabelings (Flipping Frenemies)}
\label{identify-mislabeling}\label{sec:flipping}
Based on the concept of structural balance for signed graphs, a balanced paraphrase graph can either have the entire sets of sentences being paraphrases of each other, or multiple subset groups of paraphrases with several sentences from different groups being connected with negative links. Our algorithm finds inconsistencies by identifying negative edges within a paraphrase cluster. Given the transitive relation of paraphrases, we correct the false negative links into positive. We found 88 mislabeled pairs in the QQP training set, and 2 pairs in the testing set. See Appendix~\ref{mislabeling-examples} for some examples, and~\ref{sec:conflicted-pairs} for entire list of identified pairs.

For clusters with only negative edges like the triad in Figure~\ref{weakly-balanced}, even though the relation is imbalanced according to the definition, we are unable to determine whether there should be a pair of paraphrases in the graph without knowing the actual semantic meaning of the sentences. Therefore, we use the weaker form of structural balance to represent graphs with all negative edges. We only consider the negative links within a paraphrase cluster as potentially mislabeled relations.

%% file: 4_experiments.tex
\section{Experiments}

To understand the effectiveness and impact of our augmentation and correction methods, we compare the preforms of BERT models fine-tuned to the paraphrase identification task on the original QQP dataset and three datasets derived using the graph-based methods from the previous section.

\subsection{Datasets}
\label{datasets}
The Quora Question Pairs (QQP) dataset~\citep{qqp} is based on questions extracted from Quora, where they aim to reduce the frequency of duplicated questions. Each pair is labeled as \emph{duplicate} or \emph{non-duplicate}. Duplicated questions are identified as having the same intent, meaning that they can be answered by the same answer. We consider the duplicate and non-duplicate labels comparable to paraphrase and non-paraphrase, and use the more familiar paraphrase terminology hereafter. This dataset is well suited to our approach since there are many sentences that appear in different pairs.

In addition to the original QQP dataset, we derived three additional datasets using the data augmentation and label correcting methods introduced in Section~\ref{methods}. Table~\ref{dataset-statistics} summarizes the four datasets. 

Our inference method (Section~\ref{sec:inferring}) finds over 114,000 new paraphrase pairs and 137,000 non-paraphrase pairs across the dataset, expanding the training set by over 60\%, and the testing set around 75\%. The paraphrase ratio of the augmented training set remains similar as the original set.  However, the ratio increases in the augmented testing set indicating the paraphrase clusters are sparser in the testing set. Our inconsistent label detection method (Section~\ref{sec:flipping}) detects 88 problematic labels in the training set and 2 problematic labels in the testing set. We flip the values of these labels in the \flipped{Original} and \flipped{Augmented}.\footnote{Other approaches would be worth exploring in future work such as removing the problematic pairs, manually inspecting them, and considering other labels involving sentences in problematic pairs as also likely to be problematic.}

\begin{table*}[bt]
\centering
\resizebox{\linewidth}{!}
{%
    \begin{tabular}{l c c c c c c}
    \toprule
    & \multicolumn{2}{c}{Training Set Size} & \multicolumn{2}{c}{Testing Set Size} & \multicolumn{2}{c}{Paraphrase Ratio (\%)} \\
    \multicolumn{1}{c}{Dataset} & \multicolumn{1}{c}{Paraphrase} & \multicolumn{1}{c}{Non-paraphrase} & \multicolumn{1}{c}{Paraphrase} & \multicolumn{1}{c}{Non-paraphrase} &
    \multicolumn{1}{c}{Training} &
    \multicolumn{1}{c}{Testing} \\
    \midrule
    \dataset{Original} & 134,378 & 229,468 & 14,885 & 25,545 & 36.93 & 36.82 \\
    \flipped{Original} & 134,446 & 229,380 & 14,886 & 25,544 & 36.96 & 36.82 \\
    \dataset{Augmented} & 220,890 & 363,986 & 42,570 & 28,164 & 37.77 & 60.18 \\
    \flipped{Augmented} & 220,978 & 363,898 & 42,572 & 28,162 & 37.78 & 60.19 \\
    \bottomrule
    \end{tabular}%
}
\caption{Training and testing size and percentage of paraphrase pairs for each dataset. \flipped{\emph{Dataset}} denotes the dataset with the problematic labels flipped.}
\label{dataset-statistics}
\end{table*}

\subsection{Model Training}
We fine-tune the pretrained BERT\textsubscript{BASE} model on the four datasets with the default configuration \citep{devlin-etal-2019-bert}, and implement early stopping during training. We train the model on each dataset five times independently, and report the average accuracies and F1 scores in Table~\ref{model-performance} and the detailed results with standard deviation in Appendix~\ref{sec:app-eval}.

\begin{table*}[bt]
\centering
{%
    \begin{tabular}{c l c c c c c c c c}
    \toprule
    \multicolumn{2}{r}{Testing Set:} &
    \multicolumn{2}{c}{\dataset{Original}} &
    \multicolumn{2}{c}{\flipped{Original}} &
    \multicolumn{2}{c}{\dataset{Augmented}} &
    \multicolumn{2}{c}{\flipped{Augmented}} \\
    \multicolumn{2}{c}{Model} &
    \multicolumn{1}{c}{Acc} &
    \multicolumn{1}{c}{F1} &
    \multicolumn{1}{c}{Acc} &
    \multicolumn{1}{c}{F1} &
    \multicolumn{1}{c}{Acc} &
    \multicolumn{1}{c}{F1} &
    \multicolumn{1}{c}{Acc} &
    \multicolumn{1}{c}{F1} \\
    \midrule
    
    \parbox[t]{2mm}{\multirow{3}{*}{\rotatebox[origin=c]{90}{Training Set}}} &
    \dataset{Original} & 90.35 & 87.05 & 90.09 & 86.72 & 89.72 & 91.50 & 89.72 & 91.50 \\
    & \flipped{Original} & 90.15 & 86.78 & 90.16 & 86.80 & 93.47 & 94.59 & 93.46 & 94.58 \\	
    & \dataset{Augmented} & 90.61 & 87.48 &	90.61 & 87.48 & 93.89 & 94.95 & 93.87 & 94.94\\
    & \flipped{Augmented} & \textbf{90.96} & \textbf{88.01} & \textbf{90.95} & \textbf{88.00} &\textbf{94.21} & \textbf{95.23} & \textbf{94.19} & \textbf{95.22} \\
    \bottomrule
    \end{tabular}%
}
\caption{Model performance evaluated on the four datasets. Both accuracy and F1 score are scaled by 100.}
\label{model-performance}
\end{table*}

\subsection{Result Analysis}
As shown in Table~\ref{model-performance}, the model trained on the \flipped{Augmented} dataset has the best performance (both Accuracy and F1) on all testing datasets. The improvement in model accuracy on the \dataset{Original} dataset due to augmenting the training set is modest, but significant. The improvement increases when the flipped training sets are used, and is most substantial (reaching an error rate below 6\%, compared to the original 10\% error rate) when the testing is done using the \dataset{Augmented} testing set. According to the leaderboard of GLUE benchmark~\citep{wang-etal-2018-glue}, an ALBERT based model~\citep{lan2019albert} and ERNIE~\citep{sun2019ernie} are currently the top two models on QQP task with an accuracy of 91.0\% and 90.9\% on the original testing setComparing to these state-of-the-art models, we can reach a competitive performance with the simple data augmentation proposed in this work.

The models that trained on the \dataset{Original} set has a small performance drop when tested on the \dataset{Augmented} testing set. Since this testing set has a much higher paraphrase ratio, it means that the original model is better at predicting non-paraphrases than paraphrases. It fails to give correct predictions on the augmented paraphrase pairs. This also shows the benefit of augmenting the sentence pairs by representing sentence pair relations as a graph, which helps us generate more paraphrase pairs for training and improve model accuracy on paraphrases.

Since there are only two mislabeled sentence pairs in the testing set (and 88 in the training set), it is unsurprising that the impact of flipping the inconsistent labels is small. Still, in all cases we observe the models trained with the flipped training sets have higher accuracy than those trained on the corresponding dataset with the problematic labels. Interestingly, we find that the model trained on \flipped{Original} reaches a similar performance as the model trained on \dataset{Augmented}, when tested on the \dataset{Augmented} and \flipped{Augmented} testing sets. This shows the benefits of correcting the labels identified as problematic.

%% file: 6_related.tex
\section{Related Work}
\label{sec:related}


The most closely related work from \citet{shakeel2020Paraphrase} also applies paraphrase graphs to generate additional paraphrase and non-paraphrase pairs. Similar to our method, they infer non-paraphrase pairs from sentences within different paraphrase groups, and use transitivity to find new paraphrase pairs. Different from our work, they generate additional paraphrase pairs by pairing sentences to themselves, and reversing the order of each sentence pair. Other than using structural balance, their method can only identify conflicted labels between pairs of sentences.
In addition, they only apply data augmentation on the training sets and evaluate their models directly on the original testing sets. We infer additional data and identify conflicts for both training and testing, which illustrates the full potential of our data augmentation method. 

Besides, \citet{chen-etal-2012-prefer} propose a graph-based method to improve the quality of paraphrase generation. They represent phrases as nodes and translation similarities as edges from a bilingual parallel corpus, and infer paraphrases with the pivot based method, which finds phrases with the same translation. However, this method can only infer new paraphrases within a path length of two.
\citet{Homma2017DetectingDQ} use a simpler approach by generating new paraphrase pairs with the reflexive and symmetric property of paraphrases with no graph involved. The non-paraphrase pairs are sentences randomly selected from two different pairs, which can not be guaranteed to have a correct relation.

%% file: 7_conclusion.tex
\section{Conclusion}
In this paper, we show the benefit of representing datasets as graphs. 
We develop methods based on graph theory to automatically expand a paraphrase dataset and improve labeling consistency. Our experiments show an improvement on the \flipped{Augmented} testing set after correcting the conflicted labels in the \dataset{Original} training set, and the combination of the two methods produce a model that gives the best performance across all testing sets.


%% file: appendix.tex
\appendix

\onecolumn
\section{Appendix}
\label{sec:appendix}

\subsection{Mislabeling Examples}
\label{mislabeling-examples}

\begin{figure}[htp]
\centering
\begin{subfigure}[b]{0.45\columnwidth}
\includegraphics[width=0.9\textwidth, center]{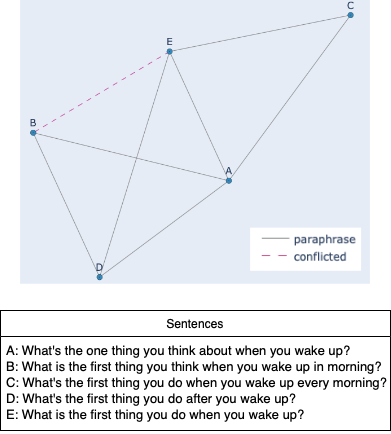} 
\caption{}
\end{subfigure}
\vspace{3mm}
\begin{subfigure}[b]{0.45\columnwidth}
\includegraphics[width=0.9\textwidth, center]{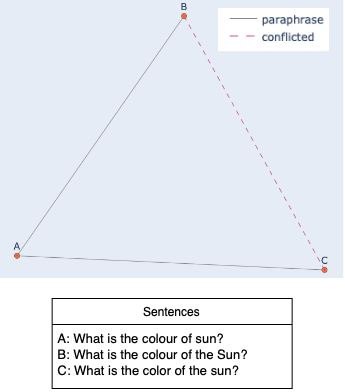} 
\caption{}
\end{subfigure}
\begin{subfigure}[b]{0.45\columnwidth}
\includegraphics[width=0.85\textwidth, center]{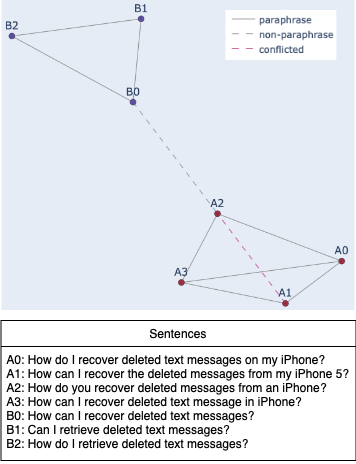} 
\caption{}
\end{subfigure}
\begin{subfigure}[b]{0.45\columnwidth}
\includegraphics[width=0.8\textwidth, center]{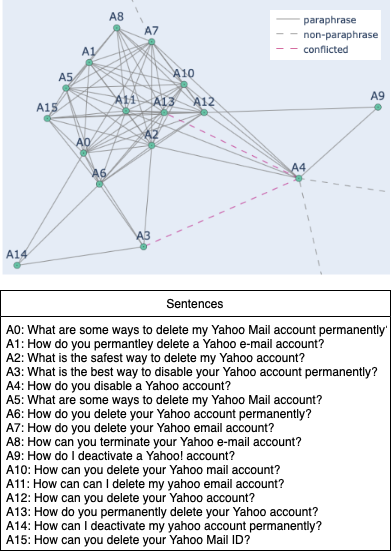} 
\caption{}
\end{subfigure}

\caption{Graphs with Inconsistent Labelings.}
\end{figure}

\clearpage

\subsection{Evaluation Results}
\label{sec:app-eval}

\begin{table*}[htp]
\centering
{%
    \begin{tabular}{c | c c c c c}
    \toprule
    & \multicolumn{5}{c}{Testing Set} \\
    \multicolumn{1}{c|}{Training Set} &
    &
    \multicolumn{1}{c}{\dataset{Original}} &
    \multicolumn{1}{c}{\flipped{Original}} &
    \multicolumn{1}{c}{\dataset{Augmented}} &
    \multicolumn{1}{c}{\flipped{Augmented}} \\
    \hline
    \multirow{3}{*}{\dataset{Original}} &   Acc & 90.35 $\pm0.14$ & 90.09 $\pm0.11$ & 89.72 $\pm0.36$ & 89.72 $\pm0.37$ \\
     & F1 & 87.05 $\pm0.33$ & 86.72 $\pm0.30$ & 91.50 $\pm 0.37$ & 91.50 $\pm0.38$ \\
     & Recall & 88.13 $\pm1.55$ & 87.92 $\pm1.53$ & 91.99 $\pm1.34$ & 91.99 $\pm1.33$ \\
     \hline
    \multirow{3}{*}{\flipped{Original}} &   Acc & 90.15 $\pm0.06$ & 90.16 $\pm0.02$ & 93.47 $\pm0.12$ & 93.46 $\pm0.12$\\
     & F1 & 86.78 $\pm0.24$ & 86.80 $\pm0.18$ & 94.59 $\pm0.11$ & 94.58 $\pm0.11$\\
     & Recall & 87.92 $\pm1.22$ & 87.96 $\pm1.13$ & 94.80 $\pm0.42$ & 94.79 $\pm0.43$\\
     \hline	
    \multirow{3}{*}{\dataset{Augmented}} &  Acc & 90.61 $\pm0.19$ & 90.61 $\pm0.21$ & 93.89 $\pm0.23$ & 93.87 $\pm0.26$\\
     & F1 & 87.48 $\pm0.24$ & 87.48 $\pm0.28$ & 94.95 $\pm0.19$ & 94.94 $\pm0.21$\\
     & Recall & 89.07 $\pm0.42$ & 89.04 $\pm0.53$ & 95.48 $\pm0.22$ & 95.47 $\pm0.23$\\
     \hline
    \multirow{3}{*}{\flipped{Augmented}} &  Acc & 90.96 $\pm0.11$ & 90.95 $\pm0.12$ & 94.21 $\pm0.04$ & 94.19 $\pm0.04$\\
     & F1 & 88.01 $\pm0.11$ & 88.00 $\pm0.12$ & 95.23 $\pm0.03$ & 95.22 $\pm0.03$\\
     & Recall & 90.14 $\pm0.40$ & 90.08 $\pm0.39$ & 96.05 $\pm0.18$ & 96.04 $\pm0.17$\\
    \bottomrule
    \end{tabular}%
}
\caption{Model performance evaluated on the four datasets. All metrics reported are scaled by 100. Standard deviations are calculated from training models five different times on the same training set.}
\end{table*}

\setlength{\belowcaptionskip}{\baselineskip}
\subsection{Sentence Pairs with Conflicted Relation}
\label{sec:conflicted-pairs}
This section shows all the sentence pairs we identified with conflicted relation in the QQP dataset. All the pairs are originally labeled as non-paraphrase, but are reachable by each other along the paraphrase links in the graph.

\begin{longtable}[h]{|c|p{14cm} c|}
    \hline No. & \multicolumn{2}{c|}{Sentence Pair} \\\hline \endhead
    \caption{Sentence pairs with conflicted relation in the QQP training set. (continued)} \endfoot
    \hline \caption{Sentence pairs with conflicted relation in the QQP training set.} \endlastfoot
    
    \csvreader[
    separator=semicolon,
    late after line=\\\hline,
    late after last line=\
    ]%
    {csv_files/mislabeled_train_pairs.csv}{question1=\qone, question2=\qtwo}%
    {\multirow{2}{*}{\thecsvrow} & \qone &\\*  & \qtwo &}%
    \label{tab:mislabeling-train}
\end{longtable}

\begin{table*}[h]
    \centering
    \begin{tabularx}{\linewidth}{|c|X c|}
        \hline No. & \multicolumn{2}{c|}{Sentence Pair} \\\hline
        \csvreader[
        separator=semicolon,
        late after line=\\\hline,
        ]%
        {csv_files/mislabeled_test_pairs.csv}{question1=\qone, question2=\qtwo}%
        {\multirow{2}{*}{\thecsvrow} & \qone &\\  & \qtwo &}%
        
    \end{tabularx}
    \caption{Sentence pairs with conflicted relation in the QQP testing set.}
    \label{tab:mislabeling-test}
\end{table*}